**Segmentation of *Drosophila* Heart in Optical Coherence Microscopy Images Using Convolutional Neural Networks**

Lian Duan,[1] Xi Qin,[1] Yuanhao He,[1] Xialin Sang,[1,2] Jinda Pan,[3] Tao Xu,[1,4] Jing Men,[5] Rudolph E. Tanzi,[6] Airong Li,[6] Yutao Ma,[4] and Chao Zhou[1,5,*]

[1]Department of Electrical and Computer Engineering, Lehigh University, Bethlehem, PA, USA
[2]Department of Electrical Engineering and Computer Science, Hainan University, Haikou, China
[3]School of Precision Instrument and Optoelectronics Engineering, Tianjin University, Tianjin, China
[4]State Key Laboratory of Software Engineering, Wuhan University, Wuhan, China
[5]Department of Bioengineering, Lehigh University, Bethlehem, PA, USA
[6]Genetics and Aging Research Unit, Department of Neurology, Massachusetts General Hospital and Harvard Medical School, USA

*Correspondence:
Chao Zhou, Department of Electrical and Computer Engineering, Lehigh University, 19 Memorial Drive West, 18015, Bethlehem, PA, USA
E-mail: chaozhou@lehigh.edu

**Abstract:**

Convolutional neural networks are powerful tools for image segmentation and classification. Here, we use this method to identify and mark the heart region of *Drosophila* at different developmental stages in the cross-sectional images acquired by a custom optical coherence microscopy (OCM) system. With our well-trained convolutional neural network model, the heart regions through multiple heartbeat cycles can be marked with an intersection over union (IOU) of ~86%. Various morphological and dynamical cardiac parameters can be quantified accurately with automatically segmented heart regions. This study demonstrates an efficient heart segmentation method to analyze OCM images of the beating heart in *Drosophila*.

## 1 INTRODUCTION

As one of the rapidly emerging imaging technologies for the biomedical research, optical coherence tomography (OCT) enables 2D cross-sectional and 3D structural imaging of tissues *in vivo* [1-3]. With high resolution and imaging speed, OCT has been widely used to provide morphological and functional information in ophthalmology, cardiology and other biomedical fields [2-4]. Optical coherence microscopy (OCM) combines the coherent detection methods of OCT and confocal detection to achieve enhanced penetration depth [5, 6]. OCM and OCT have been widely used in the field of developmental biology, including heart development of animal models such as zebrafish [7], *Drosophila* [8-12] and chick embryos [13-17] etc.

*Drosophila melanogaster*, or commonly known as fruit fly, is an important model system for developmental biology and genetic studies. There are many similarities between *Drosophila* and vertebrates in terms of early stage heart development [18]. About 75% of disease-causing genes in humans are estimated to have functional orthologs in *Drosophila* [19], and a homeobox gene [20, 21] controlling cardiac specification and morphogenesis is in both *Drosophila* and human. Thus functional analysis and genetic studies on *Drosophila* hearts can potentially apply to human heart development, disease and functional studies.

We have been using OCM to monitor the heartbeat of *Drosophila* melanogaster *in vivo* and in real time [10-12, 22, 23]. High resolution cross-section images of *Drosophila*'s heart can be obtained by OCM for heart function analysis [23]. In the meantime, automatic and high-throughput analysis is greatly desired due to the large number of specimens needed for each experiment. However, with traditional methods it is challenging to accurately and efficiently segment fly heart from OCM images. The shape of heart and the contrast of boundary vary over different developmental stages and each cardiac cycle. The fly heart chamber images become problematic especially during diastole cardiac phases, where the thinned heart wall often seems discontinuous in OCM images. Furthermore, OCM images are susceptible to speckle noises, which can cause the algorithm to misidentify the heart's boundaries [24]. Due to these factors, it is desirable to develop a reliable algorithm to segment the fly heart with high accuracy and efficiency.

Image segmentation has been a crucial problem in biomedical imaging studies. Some traditional machine learning algorithms have been implemented for this purpose, including K Means [25], random forest [26, 27], support vector machine (SVM) [28-30] and conditional random fields (CRF) [31, 32]. Yet the accuracy and universality of these algorithms are still not satisfying. However, with the advent and progress of deep learning methods, universal and powerful image classification or segmentation models can be built with proper model setup and training. In 1998, LeCun et al. [33] proposed a convolutional neural network (CNN) structure for document recognition which was followed by a successful realization by Krizhevsky et al. [34]. Since then, with the help of high performance GPU computing technique, deep learning and CNNs have become increasingly prevalent. New setups of techniques and model structures have been developed to further enhance the performance of CNNs [35-38].

Semantic segmentation is one of the rapidly developing fields, which has benefited from the development of deep learning. Long et al. [39] proposed a fully convolutional network (FCN) to accomplish semantic segmentation based on neural networks. The FCN network uses standard convolutional layers for feature extraction but substitute the last fully connected layers with a convolutional layer to generate pixel wise prediction and perform segmentation. After this study, many new segmentation network structures emerged. Instead of substituting the last fully connected layers, these new structures utilize an encoder-decoder system to perform convolution and deconvolution, and the encoder-decoder structure becomes the main structure for different semantic segmentation algorithms [40-45]. These neural network based semantic segmentation systems outperform the traditional methods in both high accuracy and speed. In this paper, we seek to implement an encoder-decoder deep learning model to perform semantic segmentation on OCM *Drosophila* heart images to segment the heart region automatically.

## 2 METHODS

### 2.1 Datasets and Training Strategy

*Drosophila* OCM images acquired in previous studies [11, 12, 23] were prepared for training and testing. The OCM system setup is shown in Figure 1. OCM images were obtained using a supercontinuum light source from Fianium with a central wavelength of ~800 nm and a bandwidth of ~220 nm, and a 2048 pixel line scan camera operating at 20k A-scans/s. A rod mirror is used to split the light into a sample arm and a reference arm. The sample arm power of the system is ~5mW and the sensitivity of the system was measured to be ~95dB. The OCM system has an axial resolution of ~1.5 um, and a transverse resolution of ~3.9 um.

For training and testing purpose, time-lapse OCM images of 90 distinct *Drosophila* were masked and used. Flies in different development stages were included to train the model. Larva, pupa and adult flies were selected for the training. Each OCM dataset contained four thousands frames of 2D fly heart images which were continuously taken in 30 seconds intervals to record the dynamic heart area. In order to increase the data variation and prevent overfitting, 100 to 500 image frames were selected out of each dataset. The selection of the images covers different shapes and sizes of the fly heart. Ground truths of all the OCM images were marked manually to indicate the region of the heart. The ground truth image is a binary image, where '1' represents the heart area and '0' represents other areas.

Data augmentation was implemented to improve the performance of the model. For each input image, operations include shifting and rotation were applied to generate additional input images, meanwhile same operation was applied on the ground truth to generate corresponding ground truth. Random shifting towards all four directions from 10 to 50 pixels were used, and rotation of 90, 180 and 270 were also used. In total 8 more augmented image ground truth pairs were generated based on one single original input image and ground truth. The augmented data were trained together with the original data. This operation would make full usage of the data. And operations like shifting would train the model with cases where the heart is not strictly in the center of the image. Data augmentation helps to train the model with different kinds of input images in order to increase the versatility of the model.

25,000 raw images before augmentation with ground truths were grouped into training, validation and testing data. In order to further examine the performance of the model, and make fully usage of the limited amount of data, a 10-fold cross-validation analysis was performed. In each round of the cross-validation, different validation and testing data were selected, and the model was trained on the rest of the data. Each round the model was trained from scratch to prevent over fitting.

The data selection in each round followed two rules. First, the data was grouped by dataset: images from the same dataset (from the same fly) were only used in training, validation or testing. Second, the selection of data in training, validation and testing covered all the developmental stages. In the beginning of each training round, ~1,000 images from 3 datasets of larva, pupa and adult flies were selected to be used as testing data, and under the same rule ~1,000 images were selected to be used as the validation data. The rest ~23,000 raw images were used as the training data, and after augmentation ~184,000 images were used for training in each round.

During the training procedure, the validation data was used to supervise the training process and the prediction accuracy on validation data was used as the trigger for early stopping mechanism. After the training was done, the model was tested using only the testing data. The usage of both validation data and testing data was to prevent the issue of over fitting caused by early stopping triggering mechanism. After each round of training, testing was made to generate an accuracy score in terms of intersection over union (IOU). After all the 10 rounds, an average IOU score can be calculated to evaluate the performance of the model.

### 2.2Network Structure

The neural network structure of our model is shown in Figure 2. The model's structure design was adopted from the U-Net design [42], and was built using Keras with Tensorflow backend. Figure 2. (a) shows the structure of the neural network model. In the figure the plates indicate the feature maps, and the numbers of feature maps are labeled. Each color arrow indicates a group of convolutional layers. In total there are five convolution groups constructing the encoder, and four deconvolution groups constructing the decoder. Image of size 128 x 128 would be the input and then a pixel based segmentation output of the same size would be generated. For the encoder component, each group contains two convolutional layers, and each of the first four groups also has a 2 x 2 max pooling layer. For the decoder component, each group contains one transpose convolutional (or deconvolutional) layer to double the feature dimensions and halve the feature channels. Each group also contains two standard convolutional layers. In addition, in each decoder group, there is one concatenate operation shown by the slim black arrow. This operation will concatenate the deconvolutional feature map with a corresponding encoder feature map. This will increase the veracity of the model and reduces information loss. For all the convolutional layers except the last one, we use a kernel size of 3 x 3 and a stride factor of 1, and Rectified linear unit (ReLU) are used as activation function for all these layers. For the last layer, an1 x 1 convolutional layer was used with sigmoid activation function to generate the predictions of being heart region for each pixel.

Adam method [46] is implemented as the optimization algorithm of the model. And intersection over union (IOU) is used as the accuracy metric to evaluate the performance of the model. IOU indicates the similarity between the ground truth and the predicted result, which is defined as:

$$IOU = \frac{PredictedResult \cap GroundTruth}{PredictedResult \cup GroundTruth}$$

A differentiable soft IOU score [47] is used as the loss function in the model. The output of the network would be differentiable instead of a standard binary output, and thus enables the back propagation inside the network to update the weights and biases. For convenience reasons, the model will also be referred as FlyNet in this paper.

## 2.3 Training and Testing

The training procedure is shown in Figure 2. (b). OCM *Drosophila* heart images in the training dataset were shuffled and then sent into the model as the input. The model then generated an output image of the segmentation result. After that the predicted segmentation (shown in blue) was compared with the ground truth image (shown in red), and loss was calculated and back propagated to update the weights and biases. This procedure was performed over many epochs until an early stopping mechanism based on validation data was triggered to prevent overfitting.

Testing procedure scheme is shown in Figure 2. (c). After the training procedure was over, a trained model was obtained. For the testing procedure the OCM images in testing group were input to the model, and the output segmentation results were generated as shown in blue color. The ground truths for the testing images were compared with the output results to evaluate the performance afterwards. Note that the ground truth of the testing result here only served for evaluation for the output results, and did not interact with the model. Once the model was well trained, it is then used to segment the fly heart region from input OCM images.

## 3 RESULTS

### 3.1 Prediction Results from the Neural Network

The model was trained on a single NvdiaGeforce GTX 1080 GPU with 8GB memory. Each epoch in the training took about 380 seconds and each round of training session stopped after about 25 epochs, when the validation results became stable and triggered the early stopping mechanism. For all the 10 rounds of training, an average IOU of ~86% was achieved. The averaged IOU is a result from testing in the 10 rounds, and has a fluctuation within 5%.

Examples of ground truth and segmentation results on larva, pupa and adult fly hearts are shown in Figure 3. Figure 3. (a, d, g) shows original *Drosophila* heart OCM images as the input for the neural network. Figure 3. (b, e, h) shows the ground truth in red color on the original OCM images. Figure 3. (c, f, i) shows the prediction output from the model in blue color on the OCM images. Different features in the images are labeled in (a)(d)(g). In larva image (a) there are tracheae (tr) on both side of the heart, while the line on the top is the cover glass / fly skin surface (cg), and the cover glass reflection artifacts (arti) are also shown in the image. In pupa and adult flies, the skin of the fly is above the heart, and there are fat tissue surrounding the heart region. In addition, videos of the masks of prediction and comparison with ground truths are available in the supplementary videos. The supplementary videos contain segmentation results for larva, pupa and adult flies, respectively (see Supporting Information Video S1, S2, S3). And the heart's beating patterns with time are shown. As can be seen, the predictions of the model are accurate and greatly resemble the shape and position of the ground truths,and the model gives accurate prediction of the heart region regardless of cardiac cycle stages or developmental stages.

### 3.2 Heart Functional Analysis

In order to quantify the performance of the FlyNet model, and compare the performance between different development stages, heart area, heart diameter and IOU data at each frame during the testing procedure were calculated. IOU is used as a key metric to examine accuracy of the model. Comparisons between ground truth and model prediction are plotted versus time (e.g. correspond to different cross-sectional image frames). Three example larva, pupa and adult fly testing datasets are picked and the heart area, heart diameter and IOU plots are shown in Figure 4. (a), (c), and (d), respectively. Areas marked manually as the ground truth and generated by the model are plotted in blue and orange color, respectively. The IOU curves of each OCM frame of the same period are plotted in green. As shown in the plots, the Drosophila's heart rate changes in different developmental stages, as shown in our previous studies [11, 12, 23]. Besides, the area curve generated by the model greatly resembles that generated by the ground truth, and the fly heartbeat can be accurately recorded. The model gives high IOU at all three different developmental stages. There are mismatches in the larva example dataset in Figure 4. (a), and the circled example region is enlarged and plotted in Figure 4. (b). These mismatches are caused by human errors and are further discussed in Discussion. Note that for adult flies, there are dips in the IOU curve, and dotted lines are plotted to indicate the periodical trend of the prediction failure in coordinate with the heart phase change. At end diastole phases and end systole phases there are notable accuracy drops, and scores for end systole phases are lower because smaller total area causes smaller denominators in the IOU equation. However, even at these points the IOU is still over 75%. Further discussion on the prediction failure causes is in Discussion section. Overall, the FlyNet model has shown good accuracy to perform the segmentation task on flies at all the developmental stages. The plots generated by neural networks exhibit clear indication of the heart beat dynamics, and numerical data and parameters can be quantified to characterize fly heart function.

To characterize the *Drosophila* heart based on OCM, it is important to analyze certain functional parameters, such as end diastolic diameter (EDD) indicating the diameter during the heart dilation, end systolic diameter (ESD) indicating the diameter during heart contraction, fraction shortening (FS) indicating the diastolic diameter lost in systole, and heart rate (HR) [8, 9]. These parameters can be calculated based on the output of the neural network. Figure 5 shows the results generated from the output of neural network, and compared with manual labeling results. The testing results from the 10-fold cross-validation were used and analyzed. In each round, the testing data contained dataset from each developmental stage, therefore overall 10 samples each from larva, pupa and adult stage were analyzed to generate the functional parameters. Figure 5 shows the comparison results between manually labeling and model predictions for ESD(a), EDD(b), FS(c), and HR(d). In each of the four figures, the respective parameters calculated by both manually labeling and model prediction are shown in purple and orange, respectively. To enhance generality, for each figure three groups of comparisons are shown indicating results for larva, pupa and adult developmental stages. Each bar in the figure represents the averaged result, and the error bar indicates the standard deviation, which shows variations of fly heart functional parameters within each group. Figure 5 shows that the model gives accurate measurements over parameters like EDD, ESD, FS and heart rate. The results show that the model is able to generate reliable data and reliably quantified parameters like EDD, ESD etc. These results prove that the neural network model is a suitable and reliable tool for the *Drosophila* heart functional analysis.

## 4 DISCUSSIONS

The convolutional neural network exhibits accuracy and stability to segment the heart of *Drosophila* in different developmental stages from OCM images. Our FlyNet model shows high extent of reliability in predicting fly heart region in diastolic, systolic phases, and in different developmental stages. Although in the OCM images there are discontinuity and noise, the deep learning model can still achieve high accuracy. In addition, based on segmentation of the fly hearts, functional parameters, such as EDD, ESD, FS and HR can be calculated with high accuracy and reliability, and match well with manual labeling and calculations.

As shown in result part and Figure 4, high IOU scores are achieved for most larva and pupa flies. In the meantime, the model's performance for some frames, especially adult fly images can be further improved. The absorption of the adult fly cuticle is high, and as a result, the adult fly images usually have lower sensitivity and sometimes the heart wall becomes less visible. In order to further examine the failure cases, examples of adult fly heart segmentation results are shown in Figure 6. Figure 6. (a, b, c) show the segmentation result for the heart in diastole phase, while Figure 6. (d, e, f) show the result for systole phase. In both the original OCM images, some part of the heart boundary is blurry and has low contrast. Figure 6. (b, e) shows the ground truth of the image, and Figure 6. (c, f) shows the prediction on the image. As can be seen from Figure 6. (c, f), due to the discontinuity of the boundary, the prediction result shown in blue has leakage on the left part of the heart, therefore causing errors in segmentation and a low IOU score. The cases shown in Figure 6 are examples of the prediction error, and many low IOU frame cases share the similar issue. Note that the error case only happens to some of the low contrast images, the well trained model can still predict accurately on many frames regardless of the vague boundary. Overall, the discontinued heart boundary in OCM images is one important factor that causes the prediction failure. In order to further enhance the performance and reduce segmentation errors, one obvious way is to improve from the data side: taking images with higher sensitivity OCM system to improve the image quality. On the other hand, increased the number of training datasets could also help improve the segmentation accuracy.

For the current model we are using images with 128 pixels in width, because the database we have from our previous research [11, 12, 23] utilized fly heart images with 128 pixels in width. Potentially images with a larger pixel size could be used to have more information, therefore enhance the accuracy of the prediction. For the work shown in this paper, however, we tried to utilize the database as is, and optimized the model to work with the current imaging setup and methods.

Another issue related to the segmentation is human error. The ground truth datasets were marked by four people manually. There are inevitable slight differences between the datasets and even within the same dataset. For example, in Figure 4. (a, b), the ground truth plot has fluctuations due to variations of manual human masking of individual OCM frames. In comparison, well-trained FlyNet model can consistently segment the fly heart from consecutive OCM frames and generate a smoother prediction line as shown in Figure 4. (a, b).

As for the model structure, we have constructed a convolutional neural network model that can be effectively trained and perform segmentation on OCM fly heart images. In order to show the high accuracy of the model and its performance, a comparison between our FlyNet model and the FCN model is made. A standard FCN-32s network [39] is built. The network contains one bilinear up-sampling layer as the decoder part. The network is trained under the same condition as our network, and achieved an average IOU score of ~65%. One training comparison example is shown in Figure 7. In the comparison example both models are trained and validated with the same datasets. As can be seen from the figure, our FlyNet model exhibits much higher IOU score and faster training speed comparing to the standard FCN network. The key reason is the decoder part of our model is designed to preserve more features to the end of the network, while the FCN network only uses limited structure in the decoder, and loses much information in the up-sampling procedure.

Another key technique to enhance the performance of the model is data augmentation. Due to the limited number of OCM images, it is important to make full usage of every image for training. The data augmentation enables the model to learn with different scenarios based on the limited images. Moreover, data augmentation prevents the model from biased predictions towards center position of the heart tube. After data augmentation, especially shifting, the model could learn to recognize the heart at other positions, and based more on the structural information, rather than the absolute position of the heart.

During the training process, we seek to build a model that can effectively differentiate the fly heart from other organs in OCM images. To make full usage of the features outside of the heart, we use OCM images containing regions with heart and other tissues, and trained a model that is capable of segmenting the heart from other organs. The other way to maintain universality is to train the three developmental stages together. Since the fly heart images differ even within the same developmental stage, it is helpful to train the model with as much diversity as possible. The combined training strategy would help enhance the model's performance.

In this paper, we present a model that utilized the 2D input images. It is relatively easy to train, and generates accurate predictions efficiently. However, the fly heart OCM data is time-lapse 3D data, and the information from sequential images is not used in our current model. The accuracy of the segmentation could be further enhanced if sequential images in the third dimension (time) are used. One possible improvement method is to construct a recurrent neural network (RNN) [48] or long short term memory (LSTM) [49] model. A RNN or LSTM model takes sequential data into account, and has good result for jobs like image captioning [50]. For semantic segmentation, there is still space for improvement on accuracy and speed. Potentially the LSTM structure could be a solution to real-time segmentation of fly heart images.

Time required to train the neural network model depends on the size of the image, and the overall size of the data. For our cases it took three to four hours to finish the training on a desktop with a single GPU. The time is relatively short, so that we can build a well performed model quickly. But once well trained the model is ready to be applied on testing data. It took only ~20 seconds to finish the segmentation for a dataset containing 4096 images. The advantage of deep learning and neural networks enables the model to be efficient in prediction, and at the same time applicable to input images from different fly developmental stages. Based on the well trained model, it is possible to build a throughput tool for heart functional analysis over OCM *Drosophila* images.

## 5 CONCLUSIONS

In summary, we utilized a convolutional neural network model to perform semantic segmentation on OCM fly heart images. Anencoder-decoder convolutional neural network was developed and trained to accurately and efficiently segment *Drosophila* heart regions from OCM image datasets.Based on the model, a pixel-wise prediction can be made and masks of the heart can be generated with an average IOU of ~86%. Furthermore, physiological parameters of the fly heartbeat, such as EDD, ESD, FS and HR, can be accurately quantified to characterize *Drosophila* heart function.

## ACKNOWLEDGEMENTS

The authors would like to thank Jason Jerwick, Luisa Göpfert and Mudabbir Badar for their help on this project. This work was supported by the Lehigh University Start-Up Fund, NSF DBI-1455613 grant, NIH K99/R00-EB010071, R15-EB019704, R21- EY026380, and R01-EB025209 grants, and National Key Basic Research Program of China 20014CB340404 grant.

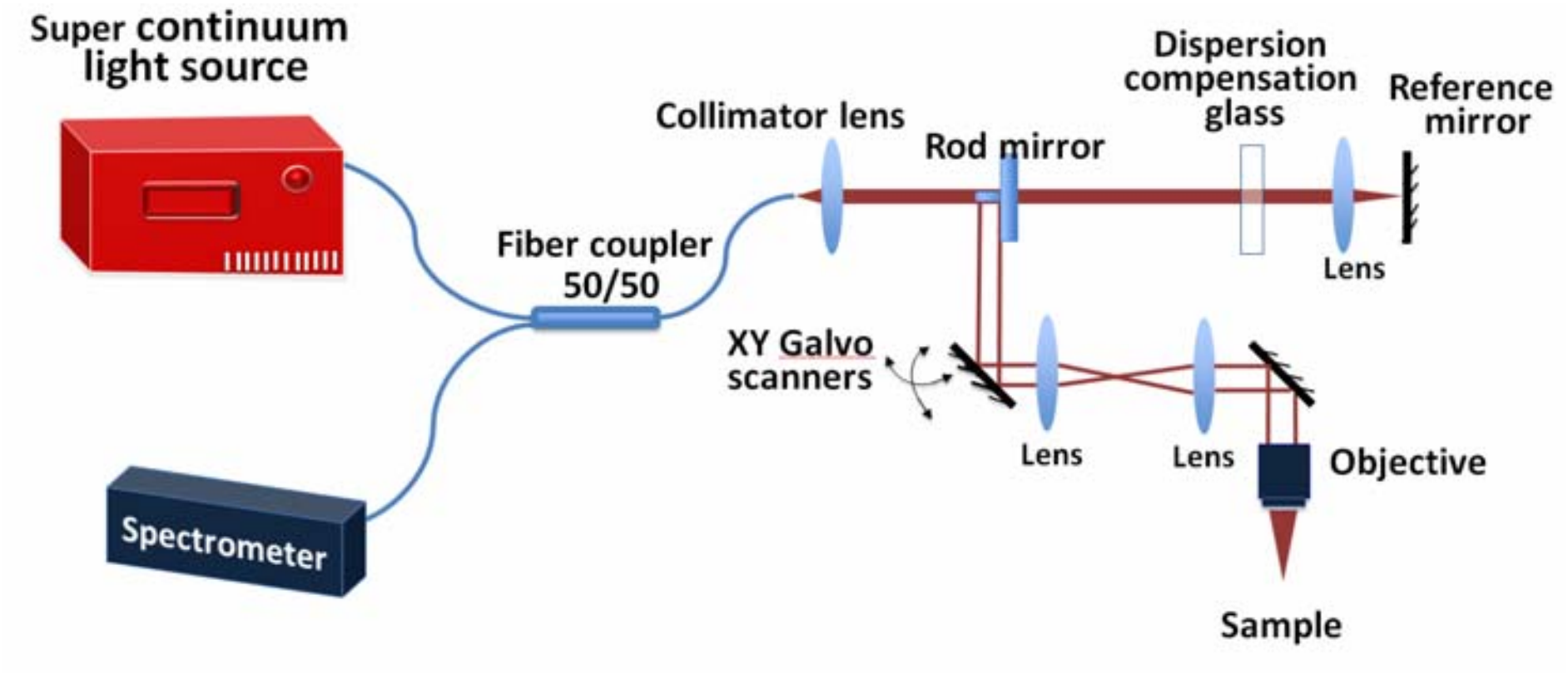

**FIGURE 1.** OCM system setup. The system uses a supercontinuum light source. A rod mirror is used to split the light into sample arm and reference arm. The back reflected light is collected by the spectrometer to generate OCM fringes.

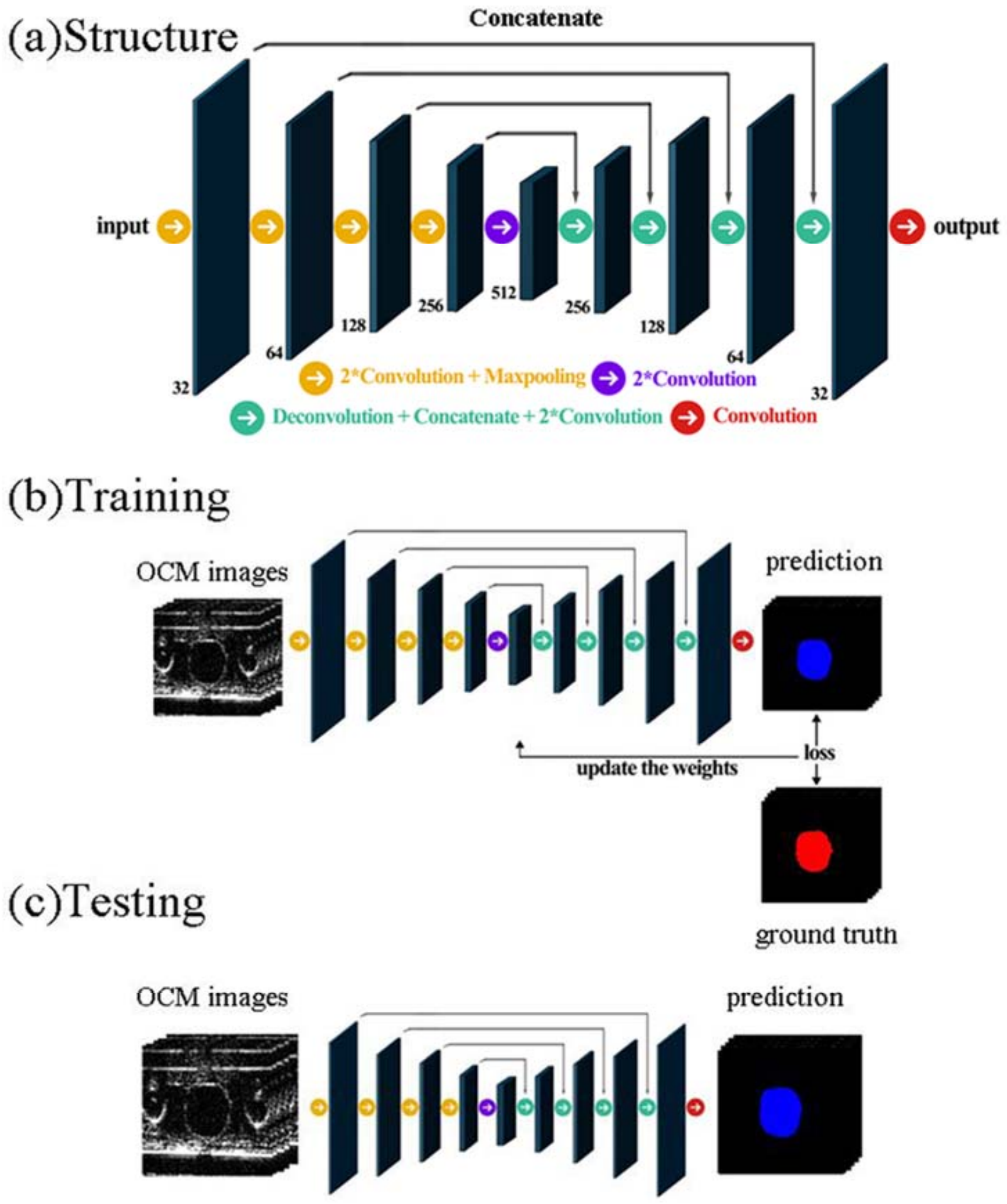

**FIGURE 2.** (a). Structure of the neural network. The feature maps are shown in green plates, with numbers by the plates indicating the number of feature maps, and the operations on feature maps are shown by colored arrows. The network consists of five convolution groups as the encoder and four deconvolution groups as the decoder. The concatenate operation is also labeled in the graph by slim black arrows. (b).Training procedure schemes. The OCM image will be input to the model and prediction of the model will be generated (shown in blue), after that manually marked ground truth (shown in red) will be introduced and compute loss function. Then the loss will back propagate into the model to update the weights using Adam's method. (c).Test procedure schemes. For testing OCM images will go into the model and predictions (shown in blue) will be generated for future analysis. No back propagation will be made in the test procedure.

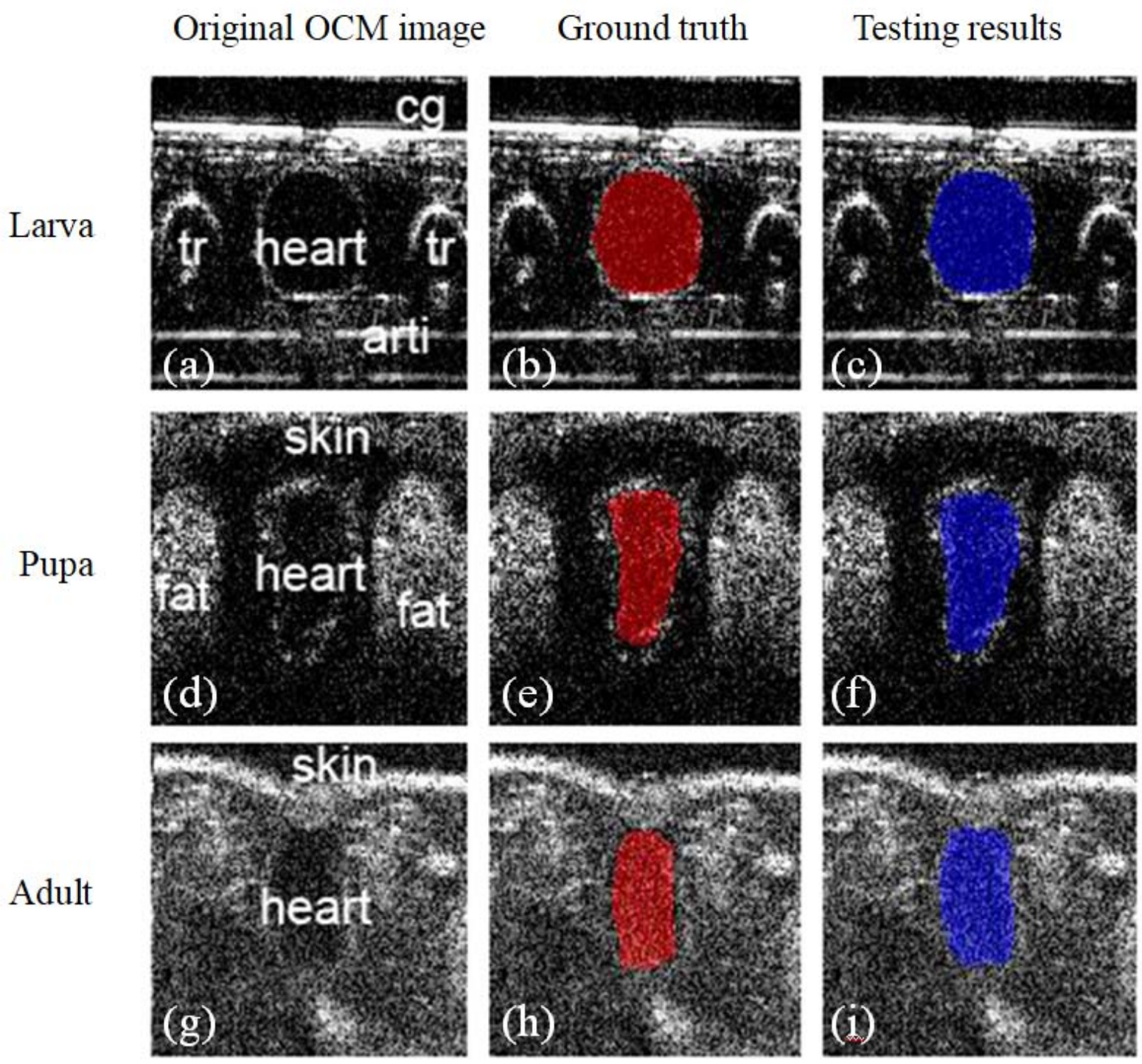

**FIGURE 3.** OCM heart image of larva (a, b, c), pupa (d, e, f) and adult (g, h, i) flies. (a, d, g) show original OCM images fly heart. Different features in the images are labeled in (a)(d)(g). The abbreviations in (a) are: tr -- tracheae, cg -- cover glass / larva skin surface, arti – cover glass reflection artifacts. The abbreviation in (b): fat – fat body. (b, e, h) show examples of ground truth labeled in red color. (c, f, i) show the corresponding testing segmentation results output from the trained model labeled by blue color. Visualization 1 shows segmentation result for larva flies, Visualization 2 shows segmentation result for pupa flies, and Visualization 3 shows segmentation result for adult flies.

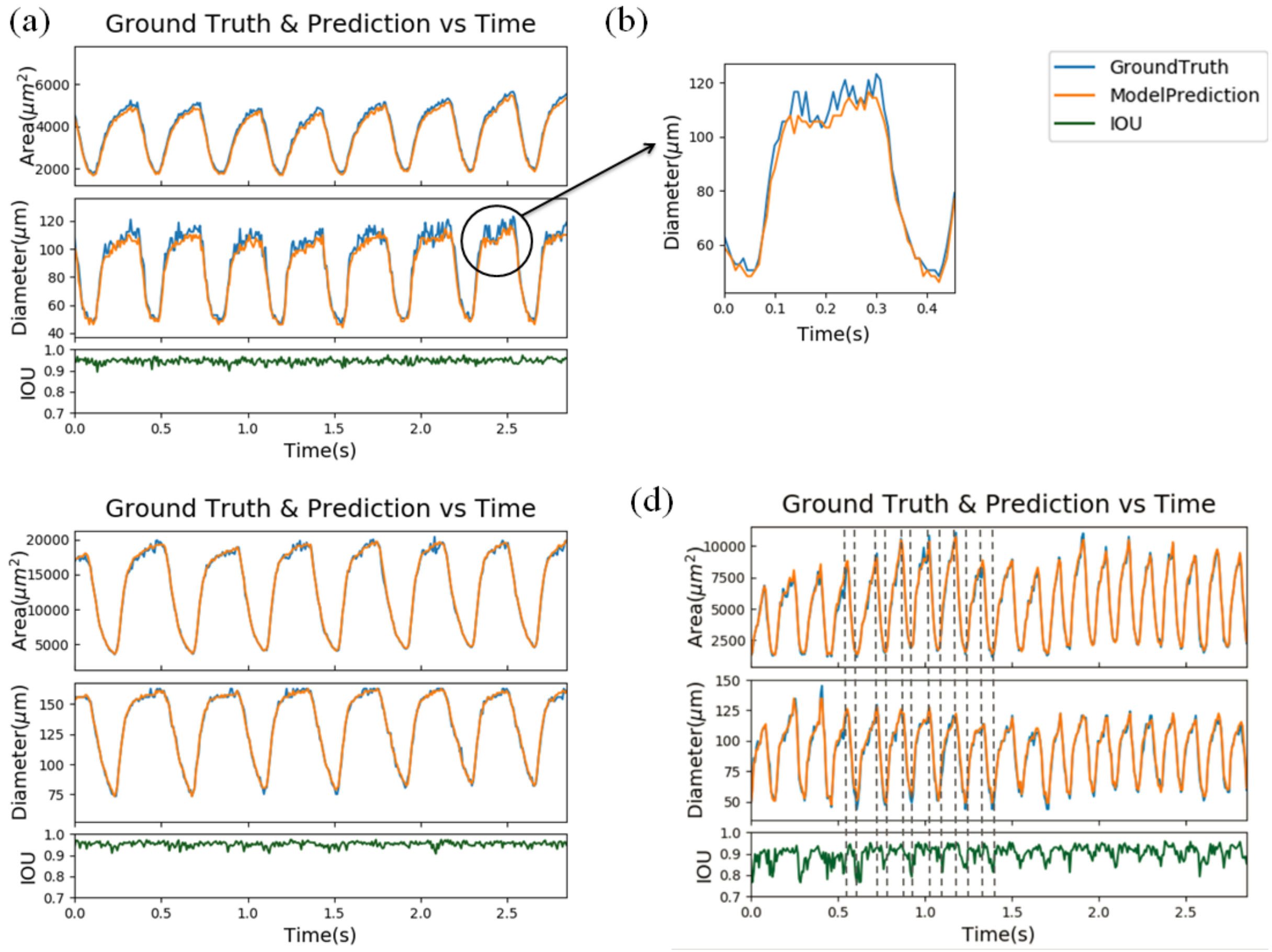

**FIGURE 4.** Heart area plots, heart diameter plots and IOU plots versus time for (a) Larva (c) pupa (d) adult flies. (b) An enlarged plot of one mismatch happens in (a). The orange plots are generated from predictions of the model, and the blue plots are generated from ground truth data. The IOU curves are shown in green. Dotted lines in (d) are plotted to indicate the trends of the heartbeat in coordinate with the low IOU period.

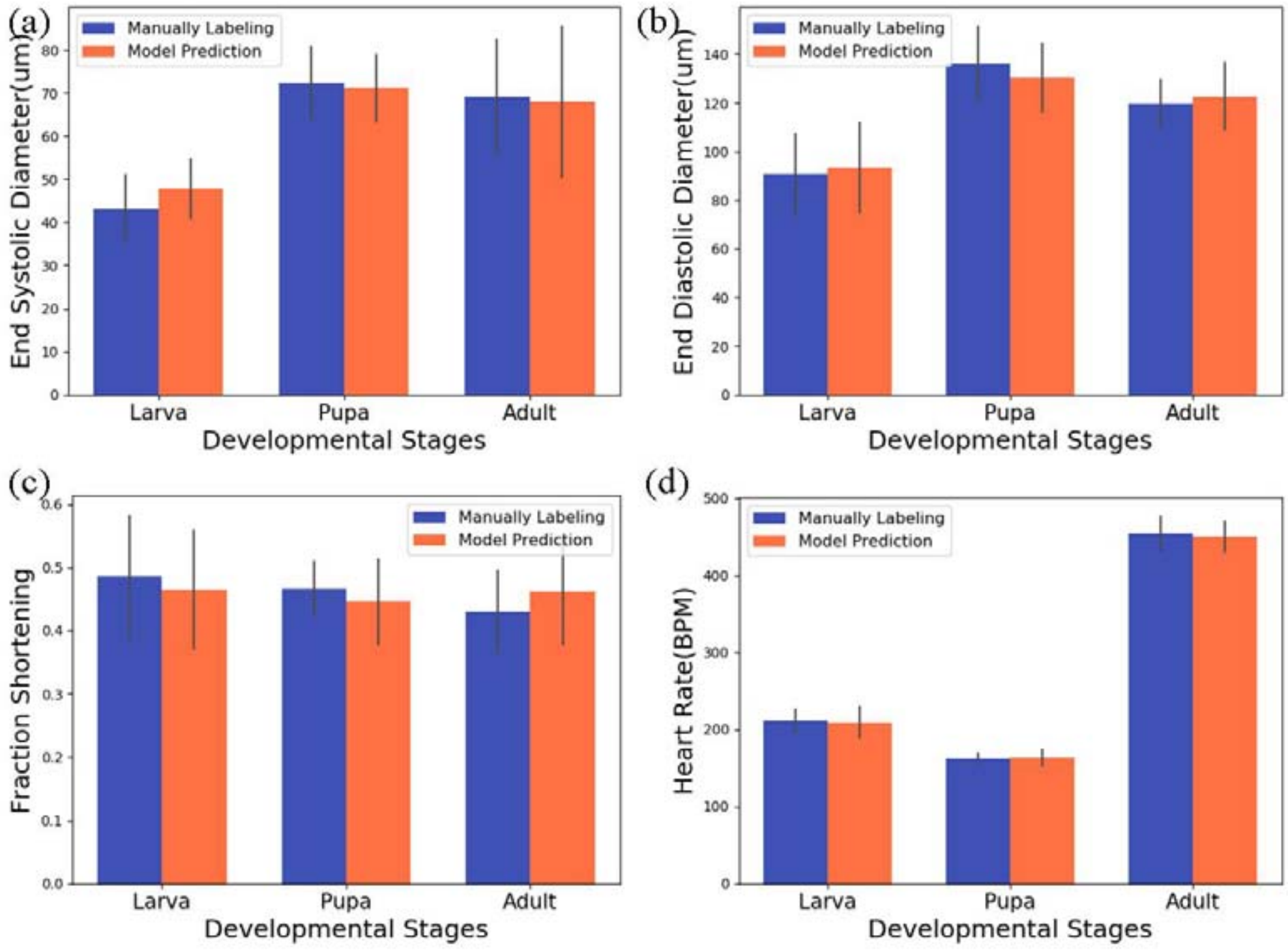

**FIGURE 5.** (a) End systolic diameter, (b) end diastolic diameter, (c) Fraction Shortening and (d) Heart Rate calculated from ground truth and prediction from the model on larva, pupa and adult flies.

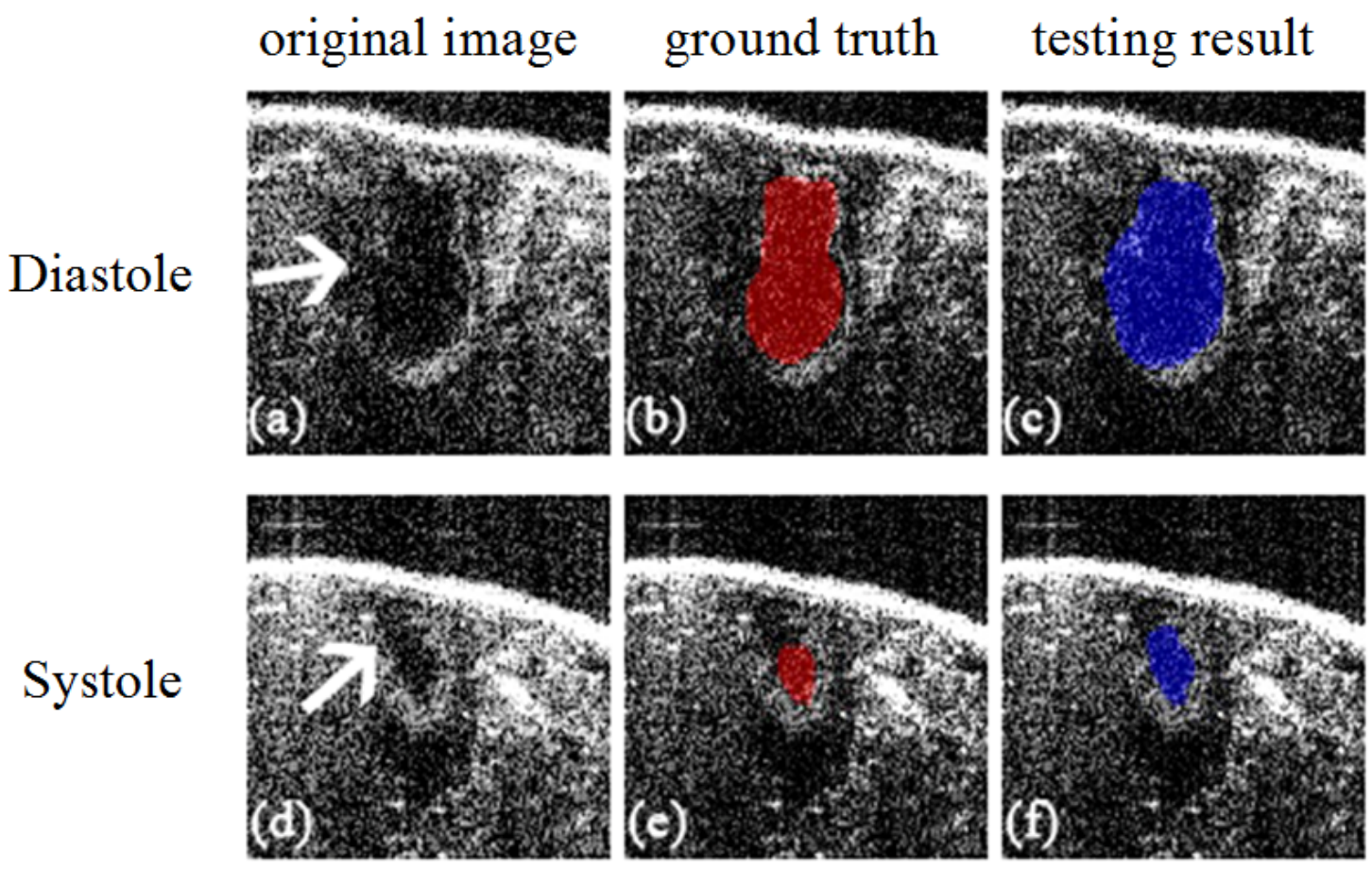

**FIGURE 6.** Segmentation error examples on an adult fly at diastole phase and systole phase. (a, d) Original image (b, e) ground truth (c, f) testing image of the OCM fly heart image. The arrow in (a) and (d) indicates the part where the boundary is blurry.

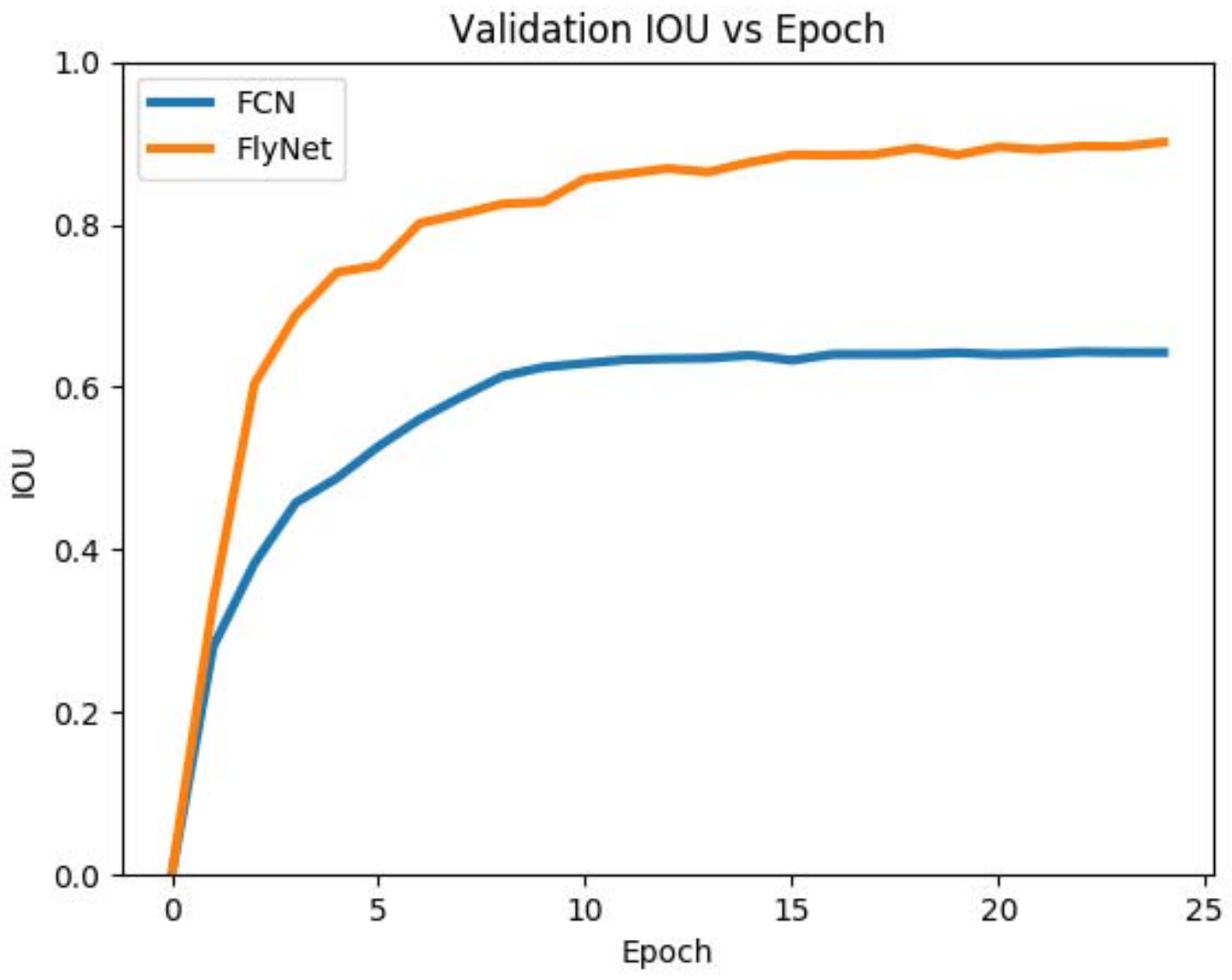

**FIGURE 7.** IOU score achieved for validation data vs epoch number during training. The blue line indicates the result obtained by a FCN-32s network, and the orange line indicates our FlyNet model's result.